\documentclass[letterpaper]{article} 
\usepackage{aaai25}  
\usepackage{times}  
\usepackage{helvet}  
\usepackage{courier}  
\usepackage[hyphens]{url}  
\usepackage{graphicx} 
\usepackage{natbib}  
\usepackage{caption} 
\usepackage{algorithm}
\usepackage{algorithmic}
\usepackage{enumitem}
\usepackage{newfloat}
\usepackage{listings}
\DeclareCaptionStyle{ruled}{labelfont=normalfont,labelsep=colon,strut=off} 
\floatstyle{ruled}
\newfloat{listing}{tb}{lst}{}
\floatname{listing}{Listing}
\usepackage{booktabs}
\usepackage{multirow}
\title{ECLAIR: Enhanced Clarification for Interactive Responses in an Enterprise AI Assistant}
\author{
    John Murzaku\thanks{Work done while interning at Adobe.}$^1$,
    Zifan Liu$^2$,
    Vaishnavi Muppala$^2$,
    \\
    Md Mehrab Tanjim$^3$,
    Xiang Chen$^3$,
    Yunyao Li$^2$
}
\affiliations{
    $^1$Stony Brook University\\ $^2$Adobe\\ $^3$Adobe Research\\
    jmurzaku@cs.stonybrook.edu, \{zifanl, mvaishna, tanjim, xiangche, yunyaol\}@adobe.com
}

\begin{document}

\maketitle

\begin{abstract}
Large language models (LLMs) have shown remarkable progress in understanding and generating natural language across various applications. However, they often struggle with resolving ambiguities in real-world, enterprise-level interactions, where context and domain-specific knowledge play a crucial role. In this demonstration, we introduce ECLAIR (\underline{\textbf{E}}nhanced \underline{\textbf{CLA}}rification for \underline{\textbf{I}}nteractive \underline{\textbf{R}}esponses), a multi-agent framework for interactive disambiguation. ECLAIR enhances ambiguous user query clarification through an interactive process where custom agents are defined, ambiguity reasoning is conducted by the agents, clarification questions are generated, and user feedback is leveraged to refine the final response. When tested on real-world customer data, ECLAIR demonstrates significant improvements in clarification question generation compared to standard few-shot methods.
\end{abstract}

\section{Introduction}
In human conversation, ambiguities are naturally detected and resolved through contextually appropriate clarification questions, integrating a combination of linguistic and paralinguistic cues. Large language models (LLMs) have made strides in this area, demonstrating some ability to resolve ambiguities through ambiguity detection and clarification question generation~\cite{kuhn2022clam,zhang2023clarify,zhang2024clamber}. However, recent LLM-based approaches tend to rely on sequential pipelines where ambiguity is first detected and then a clarification question is generated. These methods treat ambiguity detection as a purely lexical issue, ignoring vital contextual and domain-specific information. This in turn significantly limits their applicability to real-world, enterprise scenarios.

Existing approaches are also limited by their focus on datasets that lack the complexity and nuances of actual industry applications. Many of the datasets are synthetic, failing to capture the range of ambiguities in real-world interactions. This limits their applicability to enterprise systems that require a sophisticated understanding of ambiguity.

To address these challenges, we introduce ECLAIR (\underline{\textbf{E}}nhanced \underline{\textbf{CLA}}rification for \underline{\textbf{I}}nteractive \underline{\textbf{R}}esponses), a unified agentic framework designed for enterprise AI assistants. ECLAIR provides a unified conversational experience on the user side and offers better disambiguation capabilities.

Recent works on ambiguity detection and clarification question generation often use prompt-based LLM approaches, such as zero-shot or few-shot chain-of-thought (CoT) prompting \cite{kuhn2022clam,deng-etal-2023-prompting,zhang2024clamber}. Despite their successes on benchmark datasets, these approaches are inadequate for resolving real-world ambiguities in an enterprise domain setting. Our method aims to resolve these issues and is salient in two key aspects: 1) ECLAIR integrates ambiguity information and domain knowledge from multiple downstream agents, enhancing overall context-awareness in resolving ambiguities. 2) ECLAIR is developed and evaluated using a corpus of real user queries from the Adobe Experience Platform (AEP) AI Assistant \cite{bhambhri2024ai}. 

\begin{figure*}
    \centering
    \includegraphics[width=0.75\linewidth]{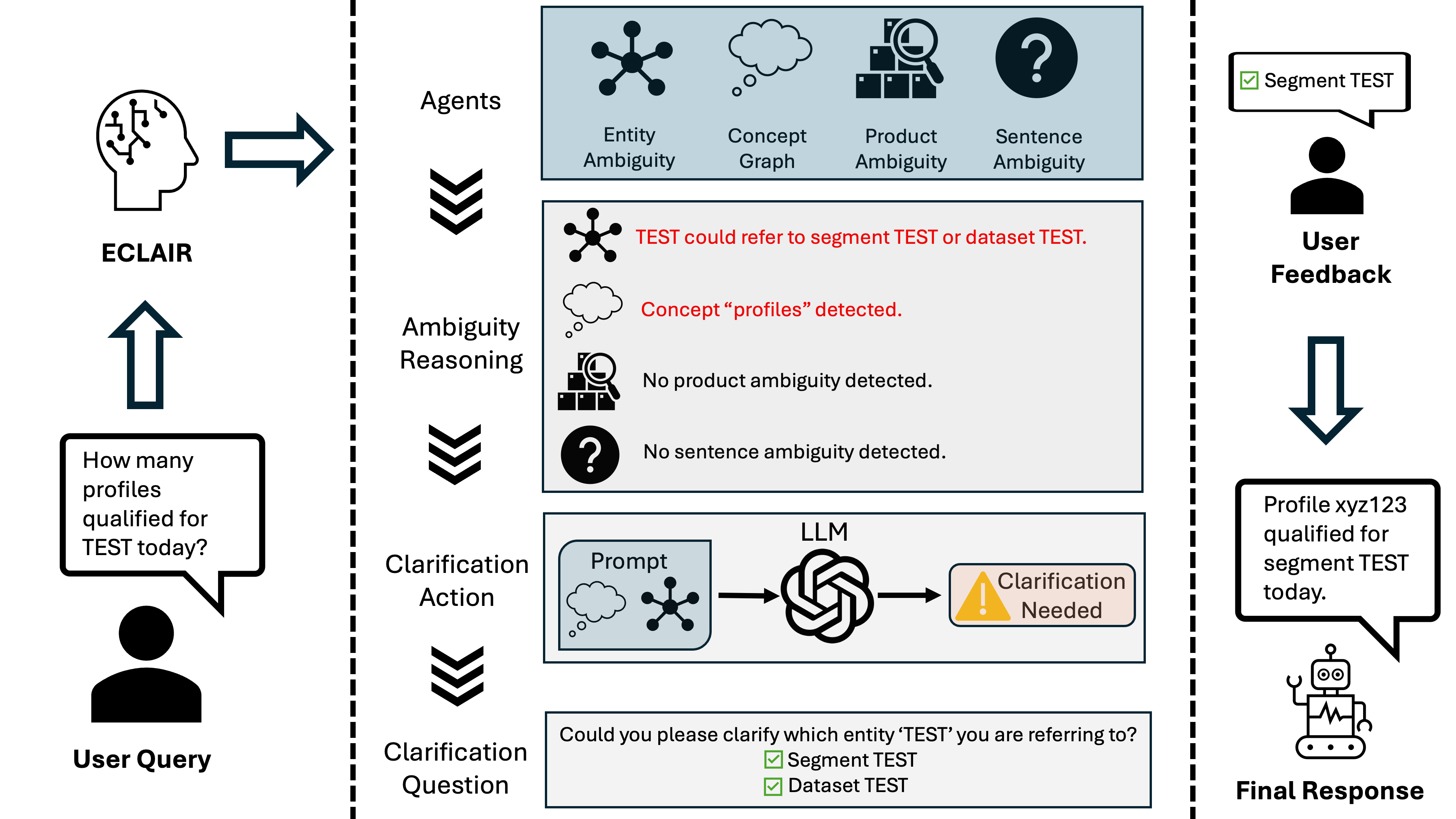}
    \caption{The architecture diagram of ECLAIR. }
    \label{fig:eclair-diagram}
\end{figure*}
\section{Framework Overview}
The ECLAIR architecture (Figure~\ref{fig:eclair-diagram}) is built as a modularizable and extendable framework: the AI Assistant developers are able to fully define their custom ambiguity detection or grounding information agents. We describe a step-by-step sample interaction and what happens in our framework.

The framework takes a user query as input and generates a set of predefined agents for ambiguity detection. The agents collaboratively reason whether an ambiguity exists in the query. The agents that do detect an ambiguity then get added to a prompt, which then calls a LLM to decide whether clarification is needed or not. In our example, ECLAIR then generates a clarification question with two choices, where the user can then add their feedback.  

In our specific use case, we define the following agents:
1) a generic sentence-level ambiguity detector that captures domain-agnostic ambiguities, 2) a product ambiguity detector that decides if the query refers to multiple products in Adobe AEP 3) an entity linking ambiguity detector that identifies text spans that can be linked to multiple entities in the database and 4) a concept graph module that identified tokens that are AEP-specific terminology.

\section{Demonstration}
Our demo video showcases two salient use cases for our enterprise AI assistant on our UI deployment. 
\paragraph{Use Case 1: Ambiguous Product} We show a case where a user's product-related question can refer to multiple Adobe products. ECLAIR successfully captures that this query is ambiguous (from the product classifier agent), and asks a relevant clarification question.

\paragraph{Use Case 2: Ambiguous Entity} We show cases where a user refers to ambiguous entities. ECLAIR successfully shows that there is entity ambiguity (through the entity linking agent), and generates a useful clarification question. The user then can clarify which specific entity type they are referring to in the interactive UI, and click their choice.

\section{Framework Evaluation}
\paragraph{Evaluation} We perform a preliminary evaluation on whether a clarification is needed or not. We evaluate this with standard Precision/Recall/F1 metrics. 

\paragraph{Dataset} Our evaluation set consists of a sample from real user queries in the AEP AI Assistant, which was annotated in-house for ambiguity detection. The dataset covers ambiguity types in three categories: contextual, where the context or reference to an object is underspecified; syntactic, where the sentence is malformed or incomplete, leading to indirect interpretation; and aleatoric, where a specific token is undefined or contains multiple possible meanings.

\paragraph{Baseline} For our baseline, we prompt GPT 3.5 to generate clarification questions if needed. 10 human-crafted examples are provided in the prompt as the few shot examples. We also use GPT 3.5 in ECLAIR for a fair comparison.

\begin{table}[h]
    \centering
    \small
    \begin{tabular}{llccc}
        \toprule
        Model & Category & P & R & F1 \\
        \midrule
        \multirow{3}{*}{Baseline} & Clar. Needed & 0.732 & \textbf{0.833} & \textbf{0.779} \\
        & Clar. Not Needed & 0.333 & 0.214 & 0.261 \\
        & Avg & 0.533 & 0.524 & 0.520 \\
        \midrule
        \multirow{3}{*}{ECLAIR} & Clar. Needed & \textbf{0.904} & 0.635 & 0.746 \\
        & Clar. Not Needed & \textbf{0.438} & \textbf{0.808} & \textbf{0.568} \\
        & Avg & \textbf{0.671} & \textbf{0.721} & \textbf{0.657} \\
        \bottomrule
    \end{tabular}
    \caption{Comparison of Precision (P), Recall (R), and F1 Score for Baseline and ECLAIR models.}
    \label{tab:comparison}
\end{table}

\paragraph{Results} Table 1 shows the results for ECLAIR compared to our few-shot baseline on deciding whether clarification question is needed or not. We first observe that ECLAIR achieves higher precision on ``Clarification Needed'', but falls short on recall and F1. We emphasize that for our production setting, we are aiming to maximize precision (that is, we do not want to over-clarify, and only ask clarifications when they are needed). Therefore, ECLAIR's higher precision means a better and more natural user experience: clarifications are only asked when necessary.
ECLAIR yields major improvements on the ``Clarification Not Needed'' label: we see a boost in precision, recall, and F1, with the largest boost occurring in recall. In turn, ECLAIR yields a overall F1 that is 13\% higher than our few-shot method. This emphasizes that ECLAIR not only captures more information and provides a clearer picture of ambiguity, but also yields better metrics compared to standard few-shot methods. 

\section{Conclusion}
We present the ECLAIR framework and demonstrate its superior performance in ambiguity detection and clarification question generation on real-world data from the Adobe Experience Platform (AEP) AI Assistant. ECLAIR's modular and unified approach not only outperforms traditional few-shot methods but also integrates seamlessly into enterprise AI systems, addressing practical challenges such as latency and precision. We showcase two major use-cases of ECLAIR in a live deployment of an enterprise AI assistant. 

\section{Acknowledgments}
We would like to thank Victor Soares Bursztyn, Cole Connelly, Nathenael Dereb, Rachel Hanessian, Jordyn Harrison, Sai Sree Harsha, Sai Jayakumar, Shun Jiang, Akash Maharaj, Danny Miller, Kun Qian, Pawan Sevak, Jordan Walker for their insightful discussions and valuable help throughout the development of this work. Their guidance and feedback greatly contributed to the quality of this research.

\bibliography{aaai25}

\end{document}